\documentclass{article} 
\usepackage{iclr2019_conference,times}

\usepackage{amsmath,amsfonts,bm}

\def\eqref#1{equation~\ref{#1}}

\def\1{\bm{1}}

\DeclareMathAlphabet{\mathsfit}{\encodingdefault}{\sfdefault}{m}{sl}
\SetMathAlphabet{\mathsfit}{bold}{\encodingdefault}{\sfdefault}{bx}{n}

\usepackage{hyperref}
\usepackage{url}
\usepackage{booktabs}
\usepackage{graphics}
\usepackage{subfig}
\usepackage{adjustbox}

\title{Advancing GraphSAGE \\ with a Data-driven Node Sampling}

\author{Jihun Oh \\
Samsung Research\\
Samsung Electronics, co., Seoul, Republic of Korea \\
\texttt{jihun2331.oh@samsung.com} \\
\And
Kyunghyun Cho \\
Department of Computer Sicence \\
New York University, New York, United States \\
\texttt{kyunghyun.cho@nyu.edu} \\
\And
Joan Bruna \\
Department of Computer Sicence \\
New York University, New York, United States \\
\texttt{bruna@cims.nyu.edu} \\
}

\iclrfinalcopy 
\begin{document}

\maketitle

\begin{abstract}
As an efficient and scalable graph neural network, GraphSAGE has enabled an inductive capability for inferring unseen nodes or graphs by aggregating subsampled local neighborhoods and by learning in a mini-batch gradient descent fashion. The neighborhood sampling used in GraphSAGE is effective in order to improve computing and memory efficiency when inferring a batch of target nodes with diverse degrees in parallel. Despite this advantage, the default uniform sampling suffers from high variance in training and inference, leading to sub-optimum accuracy.

We propose a new data-driven sampling approach to reason about the real-valued importance of a neighborhood by a non-linear regressor, and to use the value as a criterion for subsampling neighborhoods. The regressor is learned using a value-based reinforcement learning. The implied importance for each combination of vertex and neighborhood is inductively extracted from the negative classification loss output of GraphSAGE. As a result, in an inductive node classification benchmark using three datasets, our method enhanced the baseline using the uniform sampling, outperforming recent variants of a graph neural network in accuracy. 
\end{abstract}

\section{Introduction}

Machine learning on graph-structured network data has proliferated in a number of important applications. To name a few, it shows great potential in a chemical prediction problem (\cite{Gilmer2017}), a protein functions understanding, and particle physics experiments (\cite{henrion2017neural, choma2018graph}). Learning the representation of structural information about a graph discovers a mapping that embeds nodes (or sub-graphs), as points in a low-dimensional vector space. Graph neural network algorithms based on neighborhood aggregation, addressed the problem by leveraging a node's attributes (\cite{kipf2016semi, hamilton2017inductive, pham2017column}). The GraphSAGE algorithm (\cite{hamilton2017inductive}) recursively subsamples by uniform sampling a fixed number of nodes from local neighborhoods over multiple hops, and learns a set of aggregator models that aggregate the hidden features of the subsampled nodes by backtracking toward the origin. The sampling approach keeps the computational footprint of each batch in parallel computing fixed. However, despite the comprehensive features of GraphSAGE, unbiased random sampling with uniform distribution causes high variance in training and testing, which leads to suboptimal accuracy. In the present work, we propose a novel method to replace the subsampling algorithm in GraphSAGE with a data-driven sampling algorithm, trained with Reinforcement Learning. 

\section{Preliminaries: GraphSAGE}
\label{subsec:sampling_aggregation}

GraphSAGE (\cite{hamilton2017inductive}) performs local neighborhood sampling and then aggregation of generating the embeddings of the sampled nodes. The sampling step provides the benefits such that the computational and memory complexity is constant with respect to the size of a graph. Once the target node, $v\in V$, is determined, a fixed set of neighborhoods, $u^k$, is sampled as follows:
\begin{equation}
\begin{array}{l} 
    u^0 = \{ v \}, \\ 
    u^k = \cup_{\nu \in u^{k-1}} \mathcal{S}(A_{\nu}, N^k), \quad k = 1,2,...,K, 
    \label{eq:sampling-2}
\end{array}
\end{equation}
where $A_{\nu}$ is a set of neighboring nodes of $\nu$ and $N^k$ is the sample size at depth $k$. $\mathcal{S}(A_{\nu}, N^k)$ is a sampler from a uniform distribution $U(1,deg(v))$ as a default setting. This way the receptive field of a single node grows with respect to the number of layers, $K$, so the size of $\cup_{k=1}^K u^k$ is $\prod_{k=1}^K N^k$.
After the sampling, we aggregate the embeddings of nodes in the sampled set toward the original node $v$.

The initial node embeddings, $h^0_u$ for a sampled set $u$, are the input node attributes (features) $x_v$ with the dimension of $M$: 
\begin{equation}
  \begin{array}{l}
    h^0_u = x_v, \quad \forall v \in \{v\} \cup u^1 \cup \cdots \cup u^K.
  \end{array}
  \label{eq:agg-1}
\end{equation}
The mean\_concat aggregator averages the embeddings, $h^{k-1}_{\nu\in\mathcal{N}(u)}$, of the neighboring nodes, $\mathcal{N}(u)$, of a set of sampled node $u$. Then, that aggregated neighbor embedding is combined by concatenation with the embedding $h^{k-1}_u$ of a node $u$ to assign a new embedding $h^k_u$ into the node. If the concatenation is changed into the addition, it becomes the mean\_add aggregator.
\begin{align}
\text{for} \quad & k=1,2,...,K \quad \text{do}  \nonumber \\
\text{for} & \quad u \in \{v\} \cup u^1 \cup \cdots \cup u^{K-k} \quad \text{do} \nonumber \\
& h^{k}_u =\sigma \Bigg\{ \left( W_{\nu}^k \sum_{\substack{\nu \in \mathcal{N}(u)}}{\frac{h^{k-1}_{\nu}}{|\nu|}}\right) \Vert \left( W_u^k h^{k-1}_u \right) \Bigg\},
  \label{eq:mean_concat_agg}
\end{align}
where $W_{\nu}^k$ and $W_u^k$ with a size of $M'\times M$ at the first layer and $M'\times M'$ at the remaining layers are weight matrices that are shared among nodes in the network layer $k$. $M'$ is the hidden feature dimension, and $\sigma(\cdot)$ is a non-linear function, such as a rectified linear unit, defined as $\max(0,x)$. The operator $||$ indicates the concatenation of two vectors. Afterward, the new embedding, $h^k_u$, is normalized.
After finishing $K$-layer processing, the final embedding vector, $h^K_u$, is generated. This goes to a classifying layer to predict $C$-classes. The GraphSAGE model is trained to minimize classification cross-entropy loss.
\begin{equation}
  \begin{array}{l}
    L \left( \hat{y}, y \right) = -\sum_{v \in V}\sum_{\substack{i=1}}^{C} y_i \log \hat{y}_i, \quad \forall y \in Y 
  \end{array}
  \label{eq:loss}
\end{equation}

\section{Method}
\subsection{Value Function-based Reinforcement Learning for Node Sampling}
\label{subsec:value_function_RL}
To replace the previous uniform sampler, we consider a Reinforcement Learning approach which helps learning how to quickly find a good sampling distribution in a new dataset. A per-step reward, $R^k_{v,u}$, is a negative value of cross-entropy loss computed at the node $v$ given a $k$-hop uniformly subsampled neighborhood as well as a directly connected 1-hop neighborhood, $u^1$. Note that the per-step reward is a batch-wise value not applying summation over a mini-batch of target nodes, $v \in V$:
\begin{equation}
  \begin{array}{r}
        R^k_{v,u} = \sum_{\substack{i=1}}^{C} y_{v,i} \log \hat{y}^k_{v,i}=\sum_{\substack{i=1}}^{C} y_{v,i} \log \mathcal{F}_\theta(v|u^1\cup \cdots \cup u^k), \quad u \in u^1,
  \end{array}
  \label{eq:sample_func_reward}
\end{equation}
where $\mathcal{F}_\theta$ is the aggregator of GraphSAGE and inputs a target node $v$ and $k$-hop subsampled neighborhood, $u^1\cup \cdots \cup u^k$. A per-step visit count, $C^k_{v,u}$, records how many times $(v,u)$ is indexed. 
\begin{equation}
  \begin{array}{l}
        C^k_{v, u} \leftarrow C^k_{v, u} + 1, \quad u \in u^1
  \end{array}
  \label{eq:visit_count0}
\end{equation} 
The layer depth of aggregator is equal to the number of hop ($K$), as seen in the iteration count of the outer loop surrounding the aggregator (\eqref{eq:mean_concat_agg}). To produce per-step rewards, GraphSAGE predicts the classes, $\hat{y}^k$, at all the intermediate layers. To do so, we add the auxiliary classifying layers at every intermediate layer beside the final layer. We consider a return $G$ consisting of the discounted sum of per-step rewards propagated from the first hop to the final $K$-th hop:
  \begin{align}
    G_{v,u} &= R^{1}_{v,u}+\gamma R^{2}_{v,u} + ... + \gamma^{K-1} R^{K}_{v,u} = \sum^{K-1}_{k=0} \gamma^k R^{k+1}_{v,u},
  \label{eq:return}
  \end{align}
where $\gamma \in (0, 1]$ is a discount factor that discounts the contribution from the future reward. In other words, with a lower $\gamma$, we impose that a neighborhood at a closer distance has more influence on the return $G_{v,u}$. 
In order to avoid the overhead computing all the per-step rewards, we explore an approximation scheme where we set $R^k$ to zero if $k<K$. Equation \ref{eq:return} can be replaced with a version of last-hop learning approximating all-hop learning; $G_{v,u} = R^{K}_{v,u}$.
A visit count, $C_{v,u}$, sums all the per-step visit counts.
\begin{equation}
  \begin{array}{l}
        C_{v,u} = \sum_{k=1}^{K}C^k_{v,u}
  \end{array}
  \label{eq:visit_count}
\end{equation}

This return is optimized with respect to a policy $\pi$ using Reinforcement Learning. The inputs of the policy are a target node and candidates of its neighborhood, and the output action space is either $1$ or $0$ , indicating being selected as a subsample or not.
The value function associated to this policy is denoted by $\mathcal{V}_{v,u}$; we recall that it is the expected return, obtained by division of $G_{v,u}$ by $C_{v,u}$, under the policy starting from a target node $v$ to a neighboring node $u$. 
The relationship between the value function and the neighboring node $u \in u^1$ connected to the target node $v$ is defined as follows:
\begin{align}
    \mathcal{V}_{v,u} = \mathbb{E}_\pi\left[ G | S = (v,u) \right] =\frac{G_{v,u}}{C_{v,u}}, \quad \mathcal{V} \in \mathbb{R}^{|V|\times \textrm{max(deg)}}, \quad u \in u^1.
  \label{value_func}
\end{align}

\subsection{Nonlinear Regressor to Model the Value Function}
\label{subsec:nonlinear_regressor}

A possible state $(v,u)$ is not confined to a finite set of nodes observed in training. That is because it is assumed the graph is evolving; that is, unseen nodes can be observed during testing. Thus, We consider a function approximation to the value function $\mathcal{V}$ using non-linear combination of attributes at state $(v,u)$. 
\begin{equation}
  \begin{array}{r}
    \hat{\mathcal{V}}_{v,u} = \mathcal{G}_{\theta} (v, u) = 
    -\exp\left( \sigma \left( W \left( x_v || x_u \right) + b \right) \right), \quad x_v, x_u \in \mathbb{R}^{M},\quad u \in \mathcal{N}(v)
  \end{array}
  \label{nonlinear_regressor}
\end{equation}
where let $x_{v}$ and $x_{u}$ be $M$-dimensional input vectors (attributes) of a node $v$ and each member of a neighborhood, $u \in \mathcal{N}(v)$, respectively. $\theta$ denotes the weights of a differentiable non-linear regressor function, $\mathcal{G}$. A weight matrix $W$ with a size of $1\times2M$ and bias $b$ are the parameters of a single perceptron layer to be learned. This model is trained to minimize the $l2$-norm between the true value function, $\mathcal{V}_{v,u}$, obtained in \eqref{value_func} and the output, $\hat{\mathcal{V}}_{v,u}$, using mini-batch gradient descent optimization. The learned weights are shared in sampling neighborhood at all depths.   
 
\subsection{Node Sampling and Acceleration}
For subsampling a set of neighborhood $u^{k}$ of a set of node $u^{k-1}$ by reinforcement learning, we redefine the neighborhood sampling function, $S$ in \eqref{eq:sampling-2}, to include the non-linear regressor trained in subsection \ref{subsec:nonlinear_regressor}.
\begin{equation}
u^k = \cup_{\nu \in u^{k-1}} \mathcal{S}(\mathcal{G}_\theta(\nu, A_{\nu}), N^k),
\end{equation}
where $A_{\nu}$ is a set of neighboring nodes of $\nu \in u^{k-1}$. $\mathcal{G}_\theta$ is the non-linear regressor. $N^k$ is the subsample size at the $k$-th hop.
Based on the estimated value functions over the neighborhood, sorting the neighboring nodes in descending order and selecting top $N^k$ decrease the computational efficiency. To alleviate complexity and obtain the benefits of parallelism, all immediate neighbors are partitioned into $\mathcal{B}=N^k$ groups. Then, the arg$\max$ operation is executed in parallel to find the neighbor with the maximal predicted return in each batch. This scheme reduces the complexity to $O(n)$ in sequential mode or $O(n/\mathcal{B})$ in parallel mode:
\begin{equation}
  \begin{array}{l}
    \mathcal{S}(\mathcal{G}_\theta(\nu, A_{\nu}), N^k) = \{\arg_{i \in A_{\mathcal{B}(\nu)}} \max ( \mathcal{G}_{\theta}(\nu, A_{\mathcal{B}(\nu)}, N^k))\},
  \end{array}
  \label{eq:RL_sample_func_2}
\end{equation}
where let $\mathcal{B}(\nu)$ be $N^k$ groups of evenly partitioned neighborhood.

\section{Experiments}
\subsection{Experimental Setup}
For a supervised classification task on a large-scale graph in an inductive setting, we used protein-protein interaction (PPI) (\cite{zitnik2017predicting}), Reddit (\cite{Grover2016}), and PubMed (\cite{kipf2016semi}) datasets. The classification accuracy metric is a micro F1 score, combining a recall and a precision, that is commonly used in the benchmark task. We tested on the mean\_concat aggregator in \eqref{eq:mean_concat_agg} with 2 or 3 layers ($K$). The default hidden feature dimension size $M'$ is 512 in all hidden layers. The neighborhood sample size is set to 30 at all hops. We use the Adam optimizer (\cite{kingma2014adam}) and ran 10 epochs with a batch size of 32 and a learning rate of 0.01. When optimizing the non-linear regressor, we ran 50 epochs with a batch size of 512 and a learning rate of 0.001.

All the models were implemented in Python 2.7 and Tensorflow 1.12.0. Our computing environment was a single Tesla P40 GPU, 24GB memory on GPU with CUDA 9.2 and cuDNN 7.2.1 in the CentOS Linux 7. Our code can be downloaded from \url{https://github.com/oj9040/GraphSAGE_RL.git}. 

\subsection{Results}
In Table \ref{tab:RL}, the RL-based training showed over the baseline method relative improvement of 12.0\% (two-layer) and 8.5\% (three-layer) for the PPI dataset. The all-hop reward training exhibited slight superiority over last-hop reward, but the difference was not as large as the difference over the baseline. It supports the use of the last-hop approximation which is computationally more efficient. The effect of RL-based sampling was shown differently according to the distribution type and range of the observed value function. It was close to the Gaussian distribution spanned over a high and wide range for the PPI dataset while it was closely characterized by the Rayleigh distribution concentrated on a very low and narrow range for the Reddit or PubMed dataset. We can infer from the high concentration on near-zero values that the graph nodes are distributed over the relatively regular space. This may cause a marginal advantage of the RL-based sampling over the uniform sampling. 

In Table \ref{tab:cutting_edge_methods}, GraphSAGE with the RL-based sampling (*) achieved the runner-up accuracy on the PPI and the best on the Reddit and PubMed datasets. Training for longer epochs helped improving the accuracy (\#3 vs. \#5, \#4 vs. \#6). 
Beside the default GraphSAGE of mean\_concat aggregator and a sample size of [30, 30], a better compute-optimized network consisting of a mean\_add aggregator and a smaller and hop-wise decreasing sample size of [25, 10] (suggested by \cite{hamilton2017inductive}) was also performed (\#7, \#8). The parameter size of the mean\_add aggregator was approximately two third smaller than mean\_concat (refer to Par (MB)). Nevertheless, the accuracy of mean\_add was similar to or higher than mean\_concat when our proposed sampling method was applied (\#6 vs. \#8).   
The proposed method is proven to be practical and useful among these cutting-edge methods from the perspectives of high-ranked accuracy and memory and computing efficiency.

\begin{table}
  \caption{Reinforcement learning based sampling using uniform (baseline) vs. all-hop rewards vs. first-hop reward vs. last-hop reward; for parameter settings, a hidden dimension is 512, sample size is 30 for all layers, and discount rate $\gamma$ is 0.9; two or three mean$\_$concat aggregator layers plus one classification layer are constructed. Training ran for ten epochs with a batch size of 32. The shown Micro F1 score is averaged for five runs. Here, the first-hop RL is by using a very small $\gamma=0.001$.}
  \label{tab:RL}
  \centering
  \begin{tabular}{ccccc}
    \toprule
      & Uniform & All-hop RL & First-hop RL & Last-hop RL\\
      \midrule
    & \multicolumn{4}{c}{PPI} \\ 
     \cmidrule(lr){2-5}
    Two-layer ($K$=2) & 0.674 & \textbf{0.755} & 0.743 & 0.742 \\
    Three-layer ($K$=3) & 0.780 & \textbf{0.846} & 0.844 & 0.843\\
      \midrule
    & \multicolumn{4}{c}{Reddit} \\ 
     \cmidrule(lr){2-5}
    Two-layer ($K$=2) & 0.950 & \textbf{0.954} & 0.953 & 0.952\\
    Three-layer ($K$=3) & 0.959 & \textbf{0.963} & 0.961 & 0.961\\
      \midrule
    & \multicolumn{4}{c}{PubMed} \\
    \cmidrule(lr){2-5}
    Two-layer ($K$=2) & 0.879 & 0.881 & 0.882 & \textbf{0.885} \\
Three-layer ($K$=3) & 0.877 & 0.888 & 0.888 & \textbf{0.889} \\
  \bottomrule
  \end{tabular}
\end{table}

\begin{table}
  \caption{A summary of comparisons with cutting-edge methods, such as FastGCN (\cite{Chen2018}), graph attention network (\cite{velickovic2017graph}), and GraphSAGE (GS) with a mean\_concat (A1) aggregator and a sample size of [30, 30], including our proposed sampling method (*), all of which are two-layer networks. A better compute-optimized version (bold), consisting of mean\_add (A2) aggregator and the sample size of [25, 10],  is also performed. Testing time was measured in seconds for all test nodes. The shown Micro F1 score is averaged for five runs. The results from default settings, \#3 and \#4, are referred from Table \ref{tab:RL}. The `oom' indicates the runtime error due to out-of-memory.} 
  \label{tab:cutting_edge_methods}
  \centering
\begin{adjustbox}{width=1\textwidth}
  \begin{tabular}{ccccccccccc}
    \toprule
      & & \multicolumn{3}{c}{PPI} & \multicolumn{3}{c}{Reddit} & \multicolumn{3}{c}{PubMed} \\ 
     \cmidrule(lr){3-5}\cmidrule(lr){6-8}\cmidrule(lr){9-11}
      \# & Method & F1 & Time (s) & Par (MB) & F1 & Time (s) & Par (MB) & F1 & Time (s) & Par (MB) \\ 
    \midrule
    1 & FastGCN\_100ep & 0.730 & 0.10 & 1.4 & 0.945 & 0.53 & 2.4 & 0.876 & 0.040 & 0.037\\
    2 & GAT\_100ep & 0.973 & 1.29 & 12.0 &  & oom &  & 0.863 & 0.243 & 6.2\\
    3 & GS\_A1\_[30,30]\_10ep & 0.674 & 0.18 & 4.7 & 0.950 & 5.21 & 6.6 & 0.879 & 0.079 & 6.0 \\
    4 & *GS\_A1\_[30,30]\_10ep & 0.755 & 0.29 & 4.7 & 0.954 & 14.73 & 6.6 & 0.881 & 0.210 & 6.0 \\
    5 & GS\_A1\_[30,30]\_100ep & 0.746 & 0.18 & 4.7 & 0.950 & 5.21 & 6.6 & 0.888 & 0.079 & 6.0 \\
    6 & *GS\_A1\_[30,30]\_100ep & 0.785 & 0.29 & 4.7 & 0.955 & 14.73 & 6.6 & 0.890 & 0.210 & 6.0 \\
    7 & GS\_A2\_[25,10]\_100ep & 0.713 & 0.15 & 2.5 & 0.942& 2.84 & 4.5 & 0.872 & 0.023 & 4.0 \\
    8 & \textbf{*GS\_A2\_[25,10]\_100ep} & \textbf{0.813} & \textbf{0.24} & \textbf{2.5} & \textbf{0.954} & \textbf{6.99} & \textbf{4.5} & \textbf{0.898} & \textbf{0.097} & \textbf{4.0} \\
  \bottomrule
  \end{tabular}
  \end{adjustbox}
\end{table}

\section{Conclusion}
\label{sec:conclusion}
We introduced a novel data-driven neighborhood sampling approach, learned by a Reinforcement Learning, replacing random sampling with uniform distribution in GraphSAGE (\cite{hamilton2017inductive}). In order to embed nodes in a large-scale graph using limited computing and memory resources, it is crucial to sample a small set of neighboring nodes with high importance. For the supervised classification task in an inductive setting, we empirically showed that the proposed sampling method improves the node classification accuracy over the uniform sampling based GraphSAGE.

\section*{Acknowledgments}
The authors would like to thank the anonymous referees for their valuable comments and helpful suggestions. The authors collaborated with the Center for Data Science, New York University, New York, NY, USA, and were funded by Samsung Research, Samsung Electronics Co., Seoul, Republic of Korea. We express special thanks to Dr. Daehyun Kim, Dr. Myungsun Kim, and Yongwoo Lee at Samsung Research for their substantial help in supporting this collaboration.

\bibliography{GNN_DataAnalytics}
\bibliographystyle{iclr2019_conference}

\end{document}